\pdfoutput=1

\documentclass[11pt]{article}

\usepackage{acl}

\usepackage{times}
\usepackage{latexsym}
\usepackage{graphicx}

\usepackage[T1]{fontenc}

\usepackage[utf8]{inputenc}

\usepackage{microtype}

\usepackage{balance}

\balance

\title{An Evaluation Framework for Mapping News Headlines \\to Event Classes in a Knowledge Graph}

\author{Steve Fonin Mbouadeu \\
St. John's University \\
\texttt{steve.mbouadeu19@stjohns.edu}
\\\And
Martin Lorenzo  \\
IBM Research \\
\texttt{mlorenzo@ibm.com} \\\AND
Ken Barker \\
IBM Research \\
\texttt{kjbarker@us.ibm.com} \\\And
Oktie Hassanzadeh \\
IBM Research \\
\texttt{hassanzadeh@us.ibm.com} 
}

\begin{document}

%
%
%
%
%

\maketitle
\begin{abstract}
Mapping ongoing news headlines to event-related classes in a rich knowledge base can be an important component in a knowledge-based event analysis and forecasting solution. In this paper, we present a methodology for creating a benchmark dataset of news headlines mapped to event classes in Wikidata, and resources for the evaluation of methods that perform the mapping. We use the dataset to study two classes of unsupervised methods for this task: 1) adaptations of classic entity linking methods, and 2) methods that treat the problem as a zero-shot text classification problem. For the first approach, we evaluate off-the-shelf entity linking systems. For the second approach, we explore a) pre-trained natural language inference (NLI) models, and b) pre-trained large generative language models. We present the results of our evaluation, lessons learned, and directions for future work. The dataset and scripts for evaluation are made publicly available.
\end{abstract}
%
%


\section{Introduction}

Businesses and organizations can benefit from seeking knowledge of new events that may have an impact on their business.  To assist in this task, there are several media monitoring solutions with features that can provide alerts and real-time analysis for ongoing events. The majority of existing solutions are centered around entities and/or topics. For example, they identify mentions of key companies or people, group texts by topics, and analyze contents for sentiment. On the other hand, there is great value in an event-centric solution that identifies ongoing events and analyzes the characteristics of the identified events to enable event-based reasoning. In particular, such a solution would enable causal reasoning to determine the causes and consequences of ongoing events and identify potential risks and opportunities~\cite{DBLP:conf/ijcai/HassanzadehA0BB22toolkit}.

To enable a knowledge-driven event-centric news analysis and monitoring solution, a key requirement is the ability to accurately map ongoing news to event-related classes in a knowledge base. One way to perform this mapping is to treat event-related classes as a set of categories (or topics) and classify news headlines into these categories. Prior work has studied classification methods for news headlines (e.g., see~\citet{awasthy-etal-2021-ibm,ranaNewsClassificationBased2014} and references therein). The majority of existing methods rely on supervised learning and therefore require a training corpus. 
For a generic solution that can adapt to changing event classes or one that can be tuned easily for different domains, it is not feasible to rely on the availability of training corpora large enough for accurate classification.

In the absence of training data, the alternative solution is to apply unsupervised or weakly supervised classification methods that rely on little or no training data. Such methods often rely on rules and pre-trained generic models. More recently, pre-trained language models, and in particular large language models, have shown superior performance in such settings. As a result, we have seen a surge in the number of available models, each using different architectures, parameters, pre-training corpora, and fine-tuning strategies. Choosing the right model for a given task requires an evaluation framework to measure the accuracy of the models on the end task.

\begin{figure*}[t]
\begin{center}
    \includegraphics[width=0.68\textwidth]{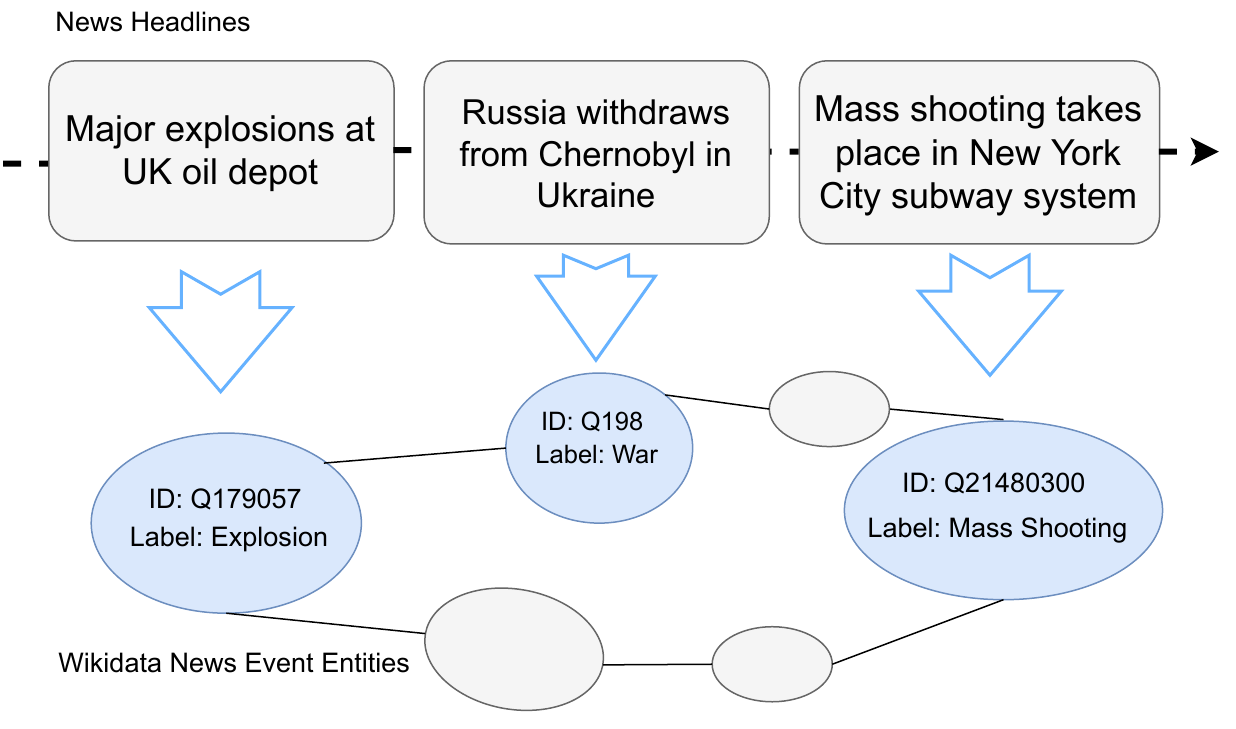}
    \caption{Example of News Headlines and Event Classes in Our Benchmark}
    \label{fig:examples}
\end{center}
\end{figure*}

In this paper, we present an evaluation framework for unsupervised mapping of news headlines to event classes in a knowledge graph. To the best of our knowledge, this is the first benchmark dataset and evaluation framework for this task. In what follows, we first present the task definition and use cases we envision for the task. We then describe our methodology for creating the benchmark dataset. Next, we present the results of our evaluation of a number of methods belonging to two different kinds of unsupervised techniques. We discuss key lessons learned and a number of avenues for future work. The datasets used in our experiments as well as the evaluation framework are publicly available~\cite{mbouadeu_steve_2023_zenodo_dataset}.



\section{Task Definition and Use Cases}

Our target task in this paper is as follows: Given a news headline and a set of event classes from a knowledge graph, find the most relevant event class to the news headline. The news headline is a short text (typically a sentence) that indicates the content of a news article by providing a concise summary of the article's contents. The knowledge graph contains event-related classes. Each class comes with one or more labels, a description of the class, and possibly a class hierarchy and other attributes. Figure~\ref{fig:examples} shows examples of news headlines, event classes, and their mappings. 
We refer to this task as {\em News Headline Event Mapping}. Note that this task is different from the event linking task defined by \citet{yuEventLinkingGrounding2023EveLINK} which takes an event mention (a phrase) and a context as input, and finds a specific Wikipedia article as output. Nevertheless, as described in Section~\ref{sec:methods}, such methods can be used for our task. 

\begin{figure*}[t]
\begin{center}
    \includegraphics[width=0.98\textwidth]{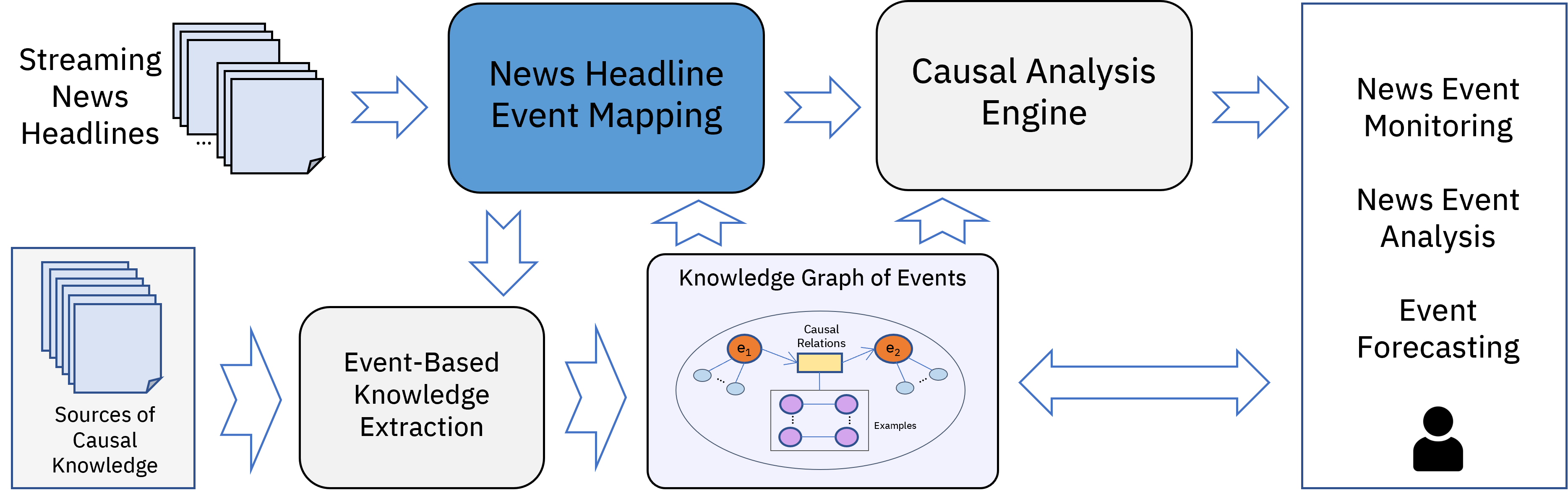}
    \caption{News Headline Event Mapping in a Knowledge-Based Event Analysis Solution~\cite{DBLP:conf/ijcai/HassanzadehA0BB22toolkit}}
    \label{fig:usecase}
\end{center}
\end{figure*}

Figure~\ref{fig:usecase} shows example use cases for news headline event mapping in the context of a knowledge-based news event analysis solution~\cite{DBLP:conf/ijcai/HassanzadehA0BB22toolkit}. In this context, news headlines from a variety of sources or a news content aggregation service (e.g., EventRegistry~\cite{eventregistry}) are monitored in order to identify major news that could have an impact on a users' organization, on a certain region, or more generally on society. This domain of interest is defined through a knowledge graph of events that contains a rich source of knowledge about past events and event classes. Such a source of knowledge can be gathered through automated knowledge extraction methods~\cite{HassanzadehBFSP20,heindorfCauseNetCausalityGraph2020} or be derived from domain-specific or general-domain knowledge sources such as Wikidata~\cite{wikidata14}. The knowledge graph provides event classes along with labels and descriptions to be used for news headline event mapping. The output of headline event mapping is then used for an analysis of the potential causes and effects of the identified event. The outcome can be used as a part of a news monitoring solution to create alerts for the identified event or its consequences so that it can assist with managing a potential risk or opportunity. It can also provide the required knowledge for an analyst looking at the implications of ongoing news for a business or organization. Finally, it can be used as an input for scenario planning~\cite{Sohrabi19} or event forecasting~\cite{embersAt4Years,RadinskyDM12_WWW}.

\section{Benchmark Dataset}
\label{sec:dataset}

\begin{figure*}[t]
\begin{center}
    \includegraphics[width=0.98\textwidth]{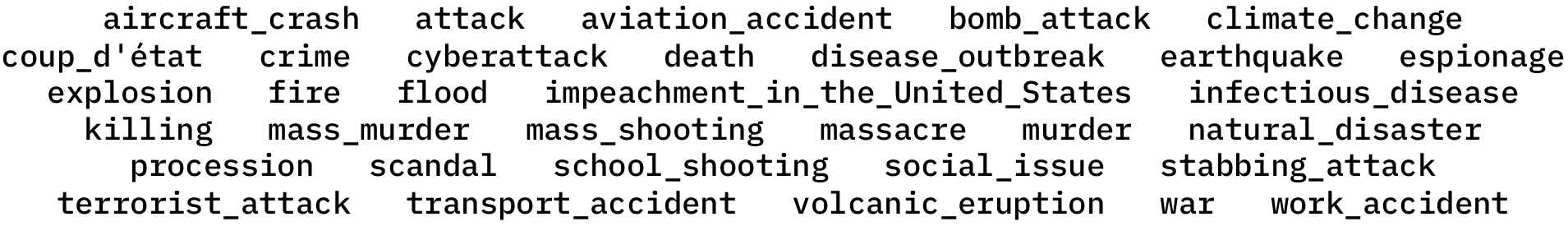}
    \caption{News Event Classes in Our Benchmark}
    \label{fig:EventClasses}
\end{center}
\end{figure*}

To the best of our knowledge, there is no benchmark dataset for the task of news headline event mapping. There are benchmarks on related tasks such as entity linking~\cite{van-erp-etal-2016-evaluating} that include news headlines. However, none of these benchmarks provide ground truth event class annotations. We have, therefore, curated a new dataset designed for the news headline event mapping task using Wikidata and Wikinews. First, we leveraged the links to Wikinews articles in Wikidata to gather a collection of event-related instances. To focus on event classes, we then filtered out instances that are not subclasses of \texttt{occurrence (Q1190554)} as well as classes with very short labels, as some non-event related entities are also linked to Wikinews. Finally, these articles were reviewed manually to check whether they related to news headlines and news events. This yielded 105 Wikinews headlines mapped to Wikidata event classes. We manually added five headlines from other sources for a final dataset of 110 mappings of headlines to Wikidata event classes. The labels of all of the Wikidata classes included in our benchmark are shown in Figure~\ref{fig:EventClasses}. The examples of Figure~\ref{fig:examples} were taken from the dataset.

There are a number of other benchmark datasets in the literature for related tasks. Closest to our task is that of zero-shot sentence classification. \citet{yin-etal-2019-benchmarking} present an excellent review of benchmarks for this task. Many benchmarks for news headline classification and for zero-shot sentence classification target binary classification (e.g., for emotions or sentiment or clickbait detection), or a small number of topics. The closest to our benchmark is the Yahoo! dataset~\cite{NIPS2015_250cf8b5yahoo}, which consists of 10 topics. To our knowledge, there is no benchmark that targets the task of assigning news headlines to event classes in a knowledge graph or a large number of well-defined topics.

\section{Evaluation}

We use our benchmark and evaluation framework to evaluate the effectiveness of a number of different kinds of methods in news headline event mapping. We first describe the approaches and implementation details for each method. We then present the results of the evaluation and a detailed discussion on key lessons learned and directions for future work.

\subsection{Methods}
\label{sec:methods}

We experiment with news headline event mapping methods ranging from a simple similarity-based baseline to adaptations of classic entity linking tools and large generative language models. 

\subsubsection{Zero-Shot Classifiers}

We evaluate two zero-shot text classification methods. One is a simple baseline based on textual similarity, while another uses state-of-the-art pre-trained language models for classification.

\paragraph{Similarity-Based Baseline} (\texttt{Fuzzy})
This method identifies a substring of a headline that is suggestive of an event occurrence, and finds the most similar event class label in the knowledge graph to that substring. The substring is found through a sliding window of bigrams and trigrams of word tokens in the input, and matching them using Levenshtein distance~\cite{10.1145/3058060.3058070} to our target event class labels. The event class with the lowest distance is returned as the most similar class to the headline.

\paragraph{Zero-Shot Text Classifier using Natrual Language Inference} (\texttt{ZSTC}) \hspace{5pt}
This classifier is a standard ``MNLI'' model: a pre-trained language model fine-tuned on the multi-genre textual entailment corpus for the Natural Language Inference (NLI) task from the RepEval workshop \cite{williams-etal-2018-broad}. Instances in the MNLI corpus are pairs of sentences with a label indicating whether the first sentence Entails the second sentence, is Contradicted by it, or is independent of it (``Neutral''). MNLI models can be used for zero-shot text classification by supplying the text to be classified as sentence1, and a textual representation of a target class as sentence2. For our experiments, the textual representation of a class (sentence2) is simply the English label of the class in Wikidata. Sentence1 is the headline to be classified. The target class whose textual representation (label) has the highest Entailment score is the predicted class for the text. Our zero-shot text classifier uses the RoBERTa-large \cite{https://doi.org/10.48550/arxiv.1907.11692} language model fine-tuned on MNLI.

\subsubsection{Classic Entity Linkers \& Adaptations}
We considered a number of state-of-the-art open-source entity linking (EL) systems to adapt to include in our experiments. Most entity linking solutions are trained to work only on named entities (e.g., people, locations, organizations) and fail when it comes to events. We considered EL systems that are more easily adaptable for mapping to event classes. The systems we considered include BLINK~\cite{wu-etal-2020-scalable}, OpenTapioca~\cite{https://doi.org/10.48550/arxiv.1904.09131}, Falcon~\cite{Sakor_2020}, and Wikifier~\cite{brank2017annotating}. Out of these, our adaptation of BLINK failed to perform well, and OpenTapioca required a training corpus. Although training OpenTapioca using our dataset provided promising results (another potential use case for our dataset), we excluded the results in this paper to focus on fully unsupervised (zero-shot) methods.

\paragraph{Falcon 2.0} (\texttt{Falcon EM})
Falcon 2.0 \cite{Sakor_2020} leverages NLP techniques to achieve state-of-the-art entity linking performance on a number of EL datasets, notably on question-structured prompts~\cite{Sakor_2020}. Given a prompt, it generates a list of entity surface forms, similar to event mentions. After generating these surface forms or tokens, it selects candidate entities for each of them by searching them in an information retrieval (IR) index (powered by \href{https://en.wikipedia.org/wiki/Elasticsearch}{Elasticsearch}) of a Wikidata data dump. We only included Wikidata concepts that were recursively instances or subclasses of event classes in the dump to tailor it to our task. 
In our evaluation, we used Falcon to match headlines to Wikidata concept labels. If Falcon did not generate at least one candidate concept, we successively stripped tokens from the right of the headline, approximating more general phrases. we repeated the process until either a candidate concept was found or the phrase became empty.
The resulting candidate concepts were then ranked using SPARQL ASK queries, measuring the taxonomic distance between the candidate concepts and our chosen news event classes. 
The class from our set of target event classes that was the shortest distance from a Falcon-generated candidate concept was chosen as the predicted class for the headline.

\paragraph{Wikifier} (\texttt{Wikifier})
\href{https://wikifier.org/}{Wikifier}~\cite{brank2017annotating} is a service for the task of ``wikification" -- taking an input text and annotating phrases in the text with Wikipedia URLs.
Wikifier employs surface forms of hyperlinks in Wikipedia to perform linking to Wikipedia entities.
For example, the Wikipedia page for earthquakes contains a link
 to the tsunami page. This suggests that earthquake is related to tsunami. 
For any surface form throughout Wikipedia that is present in the given text, Wikifier makes a candidate entity of the underlying entity. A directed mention-concept graph is created, linking surface forms to these candidate entities. Wikifier performs a global disambiguation based on the distance between entities. Distance represents the number of hyperlink hops required to get from one page to another. The smaller the distance, the more related the entities are considered. The relatedness metrics are used to score the candidate entities. Wikifier returns these candidate entities as predictions along with their scores. We converted the Wikipedia hyperlinks to Wikidata concepts with a simple lookup query. For our evaluation, we picked the top prediction that was among one of our target event classes.

\subsubsection{Large Generative Language Models} Another way to perform zero-shot classification is through the use of generative large language models (LLMs) and prompts. There are a number of LLMs available with different architectures, parameter sizes, and resource requirements. For the results in this paper, we decided to pick just one of the popular LLMs with reasonable resource requirements, namely GPT-J 6B~\cite{gpt-j}, so that our experiments are reproducible without requiring access to commercial APIs or expensive GPUs. We include two different prompting strategies for the results in this paper. Experiments with a wider variety of LLMs and more extensive prompt engineering are a subject for future work.

\paragraph{GPT-J Event Mapping} (\texttt{GPT-J EM})
Our goal here is to form a prompt that yields the generation of the relevant event class by the LLM. One way to create a prompt is to provide a few examples (a ``few-shot'' strategy) of headline + delimiter + known event class label, followed by the headline to be classified and the same delimiter, and ask the model to generate completion text. Having experimented with a number of prompting strategies, we decided to use a co-training approach \cite{pmlr-v162-lang22a}. 

Co-training works similarly to cross-validation, where each individual headline is mapped with zero shots using GPT-J and then the best-performing headlines are used to generate a few-shot prompt. The output of this method is an event label that we then mapped to Wikidata.

\paragraph{GPT-J Event Mapping with Types} (\texttt{GPT-J EMT})
We continued our experiments with GPT-J by including all the event classes in the prompt along with the pre-training. The set of labels from our news event classes were listed separately and prefixed with ``types:''. We then added this list to the beginning of the prompt to signal the categories to be picked from. We also prefixed each annotation in the pre-training examples with ``type:'' to establish that association. Additionally, we implemented a catch-all for non-event  classifications. If a prediction didn't match an event class label, we performed textual similarity matching with our target event labels to find the most similar event class to return as output.

\subsection{Results}


For our evaluation we ran each system from Section~\ref{sec:methods} on the headlines from our news event corpus to generate the systems' best predicted event classes. We calculated accuracy of each system as the percentage of top-ranked predictions matching the gold event class.

The results are shown in Table~\ref{tab:results}. In addition to the benchmark datasets, all of our outputs as well as our evaluation script are available on our GitHub repository~\cite{mbouadeu_steve_2023_zenodo_dataset}. 

\begin{table*}[t]
\begin{center}
\caption{Accuracy Results}
\label{tab:results}
\begin{tabular}{|c|c|c|c|c|c|c|c|c|c|}
\hline
\multicolumn{1}{|l|}{} & \textbf{\begin{tabular}[c]{@{}c@{}}\texttt{Fuzzy}\end{tabular}} & \textbf{\begin{tabular}[c]{@{}c@{}}\texttt{ZSTC}\end{tabular}} & \textbf{\texttt{Falcon EM}} & \textbf{\texttt{Wikifier}} & \textbf{\texttt{GPT-J EM}} & \textbf{\texttt{\begin{tabular}[c]{@{}c@{}}GPT-J EMT\end{tabular}}} \\ \hline 
\textbf{Correct @1}     & 22         & 23          & 33   & 49  & 65  & 74       \\ \hline
\textbf{Accuracy} & 0.2        & 0.209    & 0.3  & 0.445             & 0.591             & 0.673    \\ \hline
\end{tabular}
\end{center}
\end{table*}


The zero-shot classifier methods (\texttt{Fuzzy} and \texttt{ZSTC}) 
performed comparably.
They both did well on headlines that have linguistic overlap with a target class label. \texttt{Fuzzy} works when there is surface/lexical overlap, whereas \texttt{ZSTC} takes advantage of semantic overlap. Examples of headlines having linguistic overlap with target classes are: ``\textit{Major explosions at UK oil depot}'', ``\textit{Mass shooting takes place in New York City subway system}'', and ``\textit{Myanmar military vows to abide by constitution amid coup fears}''. The first two, for example, have \textit{explosion} and \textit{mass shooting} target event classes, and labels for those classes appear verbatim in the headlines. 

Linguistic overlap can result in frequent false positives, particularly for very general target classes. For example, for the headline ``\textit{More than 80 people killed in Nice, France attack on Bastille Day}'', both methods associated ``killed'' with the \texttt{killing} event class and ``attack'' with the \texttt{attack} class. Ideally, both classes would be included among the gold classes and a ranking metric used to give credit to multiple (ranked) system predictions. For simplicity, and for even comparison to systems without ranked/scored output, we only report accuracy (correct @1).


Among the classic entity linking methods, \texttt{Wikifier} performed better than \texttt{Falcon EM}.
In general, it was able to map more challenging headlines having no obvious linguistic overlap with class labels. For example, it was able to map the headline ``\textit{Russia withdraws from Chernobyl in Ukraine}'' 
to the \texttt{war} event class. 

The LLM-based methods
also showed the ability to map news headlines to event classes whose labels do not appear in the headline. Examples of such headlines are: ``\textit{Nine firefighters killed in South Carolina blaze}''
(event class \texttt{fire}),
and ``\textit{Attack at Texas elementary school kills at least 19, including 18 children}''.
(event class \texttt{school shooting}).
The second example is particularly interesting because the LLM-based methods preferred the more specific \texttt{school shooting} event class in spite of the headline's overlap with the label of the \texttt{killing} class.
The LLM-based methods (\texttt{GPT-J EM} and \texttt{GPT-J EMT}) also showed a more consistent ability to map news headlines to events with labels that are generalizations of text appearing in headlines, such as \texttt{violence} and \texttt{natural disasters}.

\subsection{Lessons Learned and Future Work}

\begin{figure}[t]
\begin{centering}
    \includegraphics[width=\columnwidth]{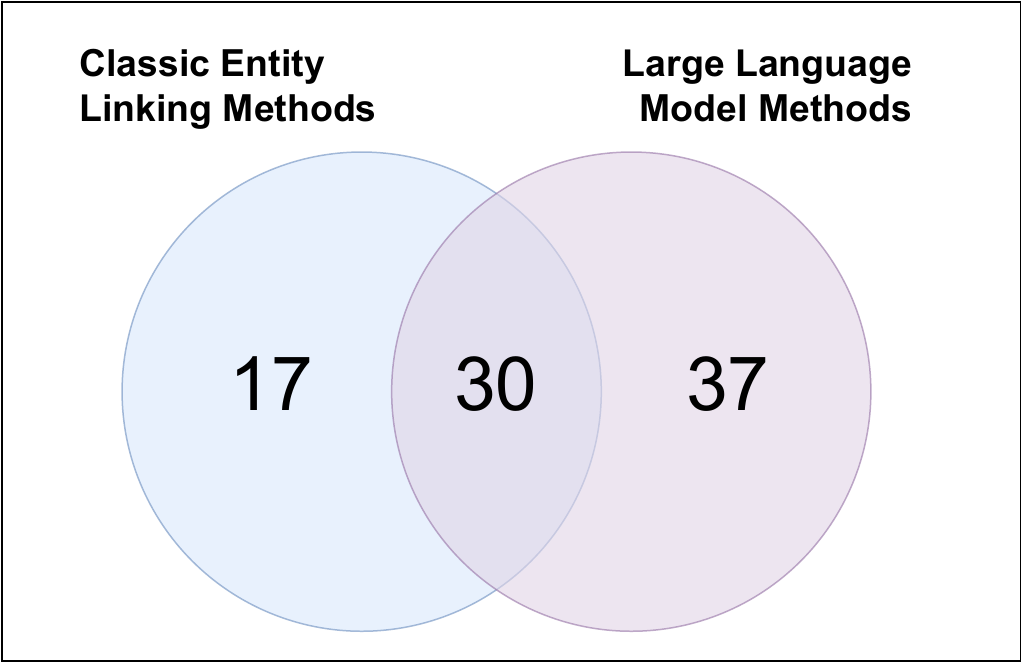}
    \caption{Overlap of Accurate Prediction Coverage of Entity Linking Adaptations and LLM-based Methods}
    \label{fig:overlap}
\end{centering}
\end{figure}

\paragraph{An Ensemble Approach}
Although the classifier and entity linking based methods did not perform as well as the LLM-based methods, they complement each other. Combining their coverage of successfully mapping headlines in the dataset yields 95\% accuracy. When comparing their coverage of the dataset, they generally succeed and fail on different types of headlines. The classic entity linker adaptations do well with headlines with single-worded event mentions that match directly to event classes. LLM-based mappers do well with multi-worded event mentions that are not necessarily substrings of the event class labels and those without any clear event mentions as well. They are still able to make the association between these more ambiguous mentions and the event entities, presumably from their learning from large amounts of text. This is further supported by the fact that of the 33\% of headlines that the LLM-based mappers failed to correctly map, 87\% have a clear event mention that closely matches the labels of their event classes. Nevertheless, there is still a noticeable amount of overlap between the two types of methods, as shown in Figure~\ref{fig:overlap}. However, these results do suggest that an ensemble approach that combines techniques used in classic entity linking and leverages large language models, intentionally deciding how and when to apply them, would improve performance on this task.

\paragraph{A Larger Dataset}
Despite the relatively small size of the current version of our dataset, we believe our results are informative, and highlight the strengths and weaknesses of different classes of methods. We also believe the small size of the data reflects well the real-world use case of building a generic and adaptable event monitoring solution, where gathering ground truth data for supervised solutions could be prohibitively expensive. Still, the methodology we outlined in Section~\ref{sec:dataset} can be extended to gather a larger and more diverse collection of news headlines mapped to event classes. At the time of writing this manuscript, we are applying a similar strategy to news headlines that are referenced from within Wikipedia-related event articles to curate a second, much larger version of our dataset.

\paragraph{More Experiments on LLMs}
With the ever-growing number of publicly-available LLMs as well as commercial APIs enabling access to such models and allowing a more extensive prompt engineering effort, our dataset and its larger extensions can be used for a study on various LLM-based news headline event mapping methods.

\section{Conclusion}

In this paper, we defined the task of news headline event mapping and outlined a few use cases for the task in event monitoring, analysis, and forecasting solutions. We presented an approach for creating a benchmark dataset, and used it to create the first benchmark dataset for the evaluation of news headline event mapping methods. We used the benchmark to evaluate different classes of mapping methods, including a) zero-short classification based methods, b) adaptations of classic entity linking methods, and c) methods based on large generative language models. Our results provide interesting insights on the strengths and weaknesses of each of the methods. We outlined several avenues for future work, including our plan to extend the dataset, work on an ensemble method, and further experiments on LLM-based methods. Our dataset, as well as our evaluation script and outputs of the models, are publicly available on our GitHub repository.


%
%
%

\bibliography{anthology,custom}
%




\end{document}